\documentclass[letterpaper, 12pt]{article}
\usepackage[left=1in, right=1in, top=1in, bottom=1in]{geometry}
\usepackage{times}
\usepackage{graphicx}
\usepackage{amsmath, amssymb}
\usepackage{cite}
\usepackage{caption}
\usepackage{titlesec}
\usepackage{booktabs}
\usepackage{float}

\titleformat{\section}{\normalfont\Large\bfseries}{\thesection.}{1em}{}
\titleformat{\subsection}{\normalfont\large\itshape}{\thesubsection.}{1em}{}
\titleformat{\subsubsection}{\normalfont\normalsize\bfseries}{\thesubsubsection.}{1em}{}

\title{Comparative Analysis of OpenAI GPT-4o and DeepSeek R1 for Scientific Text Categorization Using Prompt Engineering}

\author{
        \begin{tabular}{c}
            Aniruddha Maiti$^{1}$, Samuel Adewumi$^{1}$, Temesgen Alemayehu Tikure$^{1}$, Zichun Wang$^{1}$,\\
            Niladri Sengupta$^{2}$, Anastasiia Sukhanova$^{3}$,  Ananya Jana$^{3}$ \\
            $^{1}$Department of Mathematics, Engineering, and Computer Science, \\ West Virginia State University, Institute, WV, USA \\
            $^{2}$Fractal Analytics Inc, USA\\
            $^{3}$Department of Computer Sciences and Electrical Engineering, \\ Marshall University, Huntington, WV, USA \\
            Corresponding author: {\tt jana@marshall.edu}
        \end{tabular}
}

\begin{document}
\date{}
\maketitle

\begin{abstract}
This study examines how large language models categorize sentences from scientific papers using prompt engineering. We use two advanced web-based models, OpenAI’s GPT-4o and DeepSeek R1, to classify sentences into predefined relationship categories. DeepSeek R1 has been tested on benchmark datasets in its technical report. However, its performance in scientific text categorization remains unexplored. To address this gap, we introduce a new evaluation method designed specifically for this task. 
We also compile a dataset of cleaned scientific papers from diverse domains. This dataset provides a platform for comparing the two models. Using this dataset, we analyze their effectiveness and consistency in categorization.
\end{abstract}

\section{Introduction}

The field of artificial intelligence (AI) has advanced rapidly in recent years. These advancements have led to the development of powerful language models that are now used in many domains. Among them, OpenAI’s GPT-4o and DeepSeek R1 have received significant attention due to their capabilities and design.

OpenAI introduced GPT-4o in May 2024 \cite{openai2024gpt4o}. This model builds on earlier versions and improves performance in text, speech, and vision tasks. It provides GPT-4-level intelligence. At the same time, it operates faster and it has  lower computational cost. These improvements make GPT-4o applicable across a wide range of tasks.

DeepSeek, an AI startup, released DeepSeek R1 in January 2025 \cite{deepseek2025r1release}. This model offers performance comparable to leading proprietary systems but at a lower cost. It is open-source, which allows AI researchers to study and modify it. This transparency has made DeepSeek R1 an important contribution to the AI research community. However,  there exist concerns regarding the handling of questions related to politically sensitive topics by DeepSeek's web-based platform  \cite{guardian2025deepseek}. 

The introduction of DeepSeek R1 has influenced AI development in multiple ways. Its cost efficiency presents a challenge to companies that invest heavily in proprietary models. The open source approach has also advantageous for more research and development \cite{vox2025deepseek}.

Although DeepSeek R1 has been evaluated on general benchmarks \cite{deepseek2025technical}, its performance in the specific task of categorization of sentences in scientific text has not been analyzed in a systematic way. In particular, there has been little research on its ability to categorize scientific text. This study introduces an evaluation method designed for this task. In addition, we construct a dataset using texts from 10  scientific papers. This dataset allows a direct comparison of GPT-4o and DeepSeek R1 in sentence classification. By categorizing sentences into predefined relationship types by these two models, we examine the strengths and weaknesses. %This study provides insights into the comparative strengths and weaknesses of two recent models GPT-4o and DeepSeek R1 in scientific text categorization.

\section{Overview of GPT-4o and DeepSeek R1}

\subsection{GPT-4o: OpenAI’s Multimodal Model}
OpenAI introduced GPT-4o in May 2024 as an improvement over its previous models \cite{openai2024gpt4o}. Unlike earlier versions, GPT-4o processes text, vision, and audio together. It provides GPT-4-level intelligence but runs faster and more efficiently.

Some key features of GPT-4o are as follows:
\begin{itemize}
    \item \textbf{Multimodal Processing}: GPT-4o handles text, images, and voice inputs in a single model.
    \item \textbf{Faster Performance}: OpenAI reports that GPT-4o runs at least twice as fast as GPT-4 Turbo while maintaining similar accuracy.
    \item \textbf{Lower Cost}: The API pricing is lower than earlier versions, making it more affordable for researchers and developers.
    \item \textbf{Better Reasoning and Coding}: GPT-4o improves performance in math, programming, and long-context tasks.
\end{itemize}

OpenAI designed GPT-4o to increase efficiency and support real-time AI applications.

\subsection{DeepSeek R1: An Open-Source Alternative}
DeepSeek released DeepSeek R1 in January 2025 as an open-source large language model (LLM) \cite{deepseek2025r1release, deepseek2025technical}. Unlike many proprietary models, DeepSeek R1 is available for public use and modification. It also applies reinforcement learning to improve logical reasoning.

Some key features of DeepSeek R1 are as follows:
\begin{itemize}
    \item \textbf{Reinforcement Learning-Based Training}: The model improves its reasoning ability through reinforcement learning \cite{deepseek2025technical}.
    \item \textbf{Open-Source Access}: Multiple versions, including DeepSeek R1-Zero and smaller distilled models, are available.
    \item \textbf{Lower Computational Cost}: It performs at a level similar to OpenAI’s o1-1217 but requires fewer resources.
    \item \textbf{Community Involvement}: The open-source nature allows AI researchers to modify and build on the model.
    \item \textbf{Benchmark Performance}: The DeepSeek R1 technical report shows its competitive performance on standard reasoning and text benchmarks \cite{deepseek2025technical}.
\end{itemize}

Recent analyses have raised concerns regarding DeepSeek's web based platform's handling of questions related to politically sensitive topics \cite{guardian2025deepseek}. However,
DeepSeek R1 has contributed to making AI more accessible by providing a cost-effective, open-source model \cite{vox2025deepseek}.

\subsection{Comparative Considerations}
GPT-4o and DeepSeek R1 are designed for different users. GPT-4o focuses on real-time, multimodal applications while DeepSeek R1 provides an open-source alternative with reinforcement learning improvements. Their ability to classify scientific text has not been studied in depth.

This study introduces a method to evaluate both models for this task. We compile a dataset of 10 cleaned scientific articles and compare their sentence classification results. The next sections describe the relevant work on taxonomy of semantic relations for knowledge representation based on which the categorization is performed. Next, we briefly discuss prior work and then describe experimental design, methodology, and results.

\section{Literature Review}

The categorization of scientific text relies on the structured understanding of semantic relationships. Previous research has explored various methods for modeling these relationships to enhance AI-driven classification systems.

Maia and Lima \cite{maia2021semantic} proposed a taxonomy of semantic relations for knowledge representation, providing a foundation for categorizing scientific text. Their work highlights the importance of structuring relationships between entities to improve information retrieval and classification. Wang et al. \cite{wang2023icad} extended this idea by introducing ICAD-MI, a method for interdisciplinary concept association discovery through metaphor interpretation. Their findings underscore the significance of nuanced relationship modeling, which can be applied to AI-driven text categorization.

Corpus-based studies have also played a crucial role in understanding semantic transformations. Kunch and Kharchuk \cite{kunch2023studying} examined how deterministic vocabulary changes in artistic discourse, while Albota et al. \cite{albota2024corpus} focused on the semantic transformation of the lexeme "virus" using corpus technologies. These works highlight the dynamic nature of language and the challenges AI models face in categorizing scientific text accurately.

Additionally, Kunch et al. \cite{kunch2024studying} studied orthographic norm dynamics in media discourse, offering insights into how variations in scientific writing affect text classification. The research by Vysotska et al. \cite{vysotskainformation} on data integration in business analytics systems further demonstrates the role of information structuring in AI-driven applications.

The main idea of this study is to integrate insights from semantic taxonomies, corpus linguistics, and interdisciplinary concept modeling and use those insights in prompt engineering to examine comparative ability of two large language models, GPT-4o and DeepSeek R1 in the context of processing and categorizing scientific text .

\section{Methodology}
\subsection{Data Collection}

To evaluate the performance of GPT-4o and DeepSeek R1 in scientific text categorization, we curated a dataset consisting of ten scientific articles sourced from arXiv. These articles were selected to ensure broad disciplinary coverage. Articles from the following categories are used:

\begin{itemize}
    \item \textbf{Physics}: Astrophysics, Condensed Matter Physics
    \item \textbf{Computer Science}: Databases, Networking and Internet Architecture, Systems and Control
    \item \textbf{Economics}: Theoretical Economics, General Economics
    \item \textbf{Life Sciences}: Biomolecules, Neural and Evolutionary Computing
    \item \textbf{Engineering}: Signal Processing
\end{itemize}

While large-scale datasets with thousands of papers exist, we selected a smaller, carefully curated set of ten papers for several reasons. First, each scientific paper contains substantial textual data, typically spanning multiple sections such as introductions, methodologies, discussions, and conclusions. This provided us with a sufficient number of sentences to conduct meaningful comparative analysis without excessive redundancy. Second, by maintaining a manageable dataset size, we were able to manually verify extracted sentences for accuracy, at the same time maintaining the integrity of the evaluation. Lastly, selecting papers across different domains allowed us to assess how well each model generalizes to different styles of scientific writing.

The constructed dataset serves as a benchmark for evaluating sentence categorization based on predefined relationship categories, enabling a structured comparison between GPT-4o and DeepSeek R1.

\subsection{Text Preprocessing}

The extracted text undergoes preprocessing to remove equations, footnotes, and inline citations. The primary goal is to ensure clean, structured input for categorization while preserving the original paragraph structure. Instead of processing sentences individually, which would introduce significant computational and cost (in terms of tokens) overhead, we process one paragraph at a time to optimize efficiency.

Each paragraph is formatted as follows:
\begin{itemize}
    \item \textbf{Equations and Mathematical Notation Removal}: Mathematical expressions, LaTeX-style inline formulas, and standalone equations are filtered out to prevent interference with textual analysis.
    \item \textbf{Citation and Footnote Removal}: We wanted to use clean and grammatically complete sentences. For this reason, citations (such as “[12]”, “(Smith et al., 2020)”) and footnotes are stripped so that they can not interfere with the categorization process.
    \item \textbf{Sentence-Level Structuring}: Although categorization was performed on sentence level, while processing using structured prompts, we processed one paragraph at a time. Sentences within each paragraph were kept intact. This was required for logical flow and contextual coherence.
    \item \textbf{Blank Line Separation}: Paragraphs are separated by blank lines to facilitate structured input handling.
\end{itemize}

This paragraph-level preprocessing method is adopted for designing an efficient prompt structure. Since every prompt must contain the full list of 17 relationship categories along with examples, processing at the paragraph level significantly reduces the total number of API calls while maintaining accuracy. This approach also helps retaining full contextual meaning.

\subsection{Categorization and Prompt Engineering}
Scientific writing often contains complex sentence structures. This makes the preparation of relationship category list a challenging task. We developed a total of 17 relationship categories. The principles we followed while developing the list of categories are:
\begin{itemize}
    \item \textbf{Merging Overlapping Categories}: Through merging and consolidation of similar categories we reduce the chance of overlapping so that it reduces redundancy between categories with similar meanings. It simplifies classification making relationships easier to use and gives rise to more logical grouping improving clarity along with accuracy. 
    \item \textbf{Eliminating Highly Specific Categories}:  categories  too specific and were absorbed into broader, more versatile categories in the process such that it prevents unnecessary complexity, encourages generalization making categories more widely applicable and prevents tangling with other relationships that already express similar concepts.
    \item \textbf{Improving Naming for Clarity and Accessibility}: We simplify category names so that it is easier to understand, is applicable across different fields and is more intuitive.
    
    \item \textbf{Improving Scope and Applicability}: The new categories are designed for broader applicability in scientific, technical, and research-based publications such that it reduces ambiguity when applying relationships to complex texts.
    \item \textbf{Logical and Conceptual Streamlining}: We improve logical structuring so that each category is conceptually distinct from the others. The improved logical structuring minimizie confusion with different logical categories.
\end{itemize}

The categorization process assigns each extracted sentence to one of 17 predefined relationship categories. These categories include Part-Whole, Cause-Effect, Interaction, Comparison, and Time-Based relationships. 
Instead of processing sentences individually, we classify one paragraph at a time. This preserves contextual meaning and reduces computational overhead. Processing full paragraphs allows a structured approach while keeping relationships intact.

\subsubsection{Relationship Categories}
The 17 relationship categories define different types of connections between entities. Each category describes how two entities (A and B) interact. Below is an overview:

\begin{itemize}
    \item \textbf{Part-Whole Relationship}: A is a part of B or contains B.  
    Example: "A mitochondrion is part of a cell."

    \item \textbf{Category \& Type Relationship}: A is a specific instance of category B.  
    Example: "A rose is a type of flower."

    \item \textbf{Cause \& Effect Relationship}: A causes or leads to B.  
    Example: "Smoking causes lung cancer."

    \item \textbf{Condition \& Rule Relationship}: If A happens, B follows.  
    Example: "If water reaches 100°C, it boils."

    \item \textbf{Action \& Change Relationship}: A changes or transforms B.  
    Example: "Heating metal expands it."

    \item \textbf{Interaction \& Influence Relationship}: A and B influence each other.  
    Example: "Gut bacteria influence human metabolism."

    \item \textbf{Comparison Relationship}: A is similar to or different from B.  
    Example: "Electric cars are more efficient than gasoline cars."

    \item \textbf{Opposing Relationship}: A prevents or contradicts B.  
    Example: "Vaccination prevents disease."

    \item \textbf{Time-Based Relationship}: A happens before or after B.  
    Example: "The Renaissance happened before the Industrial Revolution."

    \item \textbf{Location-Based Relationship}: A is inside, near, or above B.  
    Example: "The nucleus is inside the cell."

    \item \textbf{Quantity \& Measurement Relationship}: A is greater than or proportional to B.  
    Example: "Speed is proportional to distance over time."

    \item \textbf{Ownership \& Control Relationship}: A owns or controls B.  
    Example: "A company owns patents."

    \item \textbf{Limitation \& Restriction Relationship}: A limits or stops B.  
    Example: "Budget constraints limit research progress."

    \item \textbf{Representation \& Symbol Relationship}: A represents or encodes B.  
    Example: "DNA encodes genetic information."

    \item \textbf{Replacement \& Substitution Relationship}: A replaces or is equivalent to B.  
    Example: "Solar energy replaces fossil fuels."

    \item \textbf{Formation \& Emergence Relationship}: A emerges from B or leads to the formation of B.  
    Example: "Planets form from cosmic dust."

    \item \textbf{Process \& Change Over Time Relationship}: A transitions into B.  
    Example: "A caterpillar turns into a butterfly."
\end{itemize}

\subsubsection{Prompt Engineering Strategy}
To categorize sentences efficiently, we use a structured prompt that processes one paragraph at a time. The prompt follows a fixed structure for consistency. Processing at the paragraph level reduces redundancy and improves efficiency.

The prompt includes:
\begin{itemize}
    \item A short introduction explaining the task.
    \item The list of all 17 relationship categories with definitions and examples.
    \item The paragraph extracted from the dataset.
    \item A request for the model to classify each sentence and extract entities A and B.
\end{itemize}

An example prompt format is shown below:

\begin{verbatim}
You will be given a paragraph from a scientific paper. Your task
is to categorize each sentence within the paragraph into one of 
the 17 predefined relationship categories listed below. For each
sentence, extract the two primary entities  (A and B) involved in
the relationship. The possible relationship categories are:

1. Part-Whole Relationship (A is part of B, A contains B)
   Example: "A mitochondrion is part of a cell."

2. Category & Type Relationship (A is a type of B, A belongs
to category B)
   Example: "Gravity is a fundamental force in physics."

... (remaining categories) ...

Now, classify the following paragraph:
<A PARAGRAPH FROM ARTICLE IS INSERTED HERE>

Provide output in the following format:
Sentence: <Extracted sentence>
Category: <Selected category>
A: <Entity A>
B: <Entity B>
\end{verbatim}

\subsubsection{Refinement and Validation}
AI models may misclassify relationships. To improve accuracy, we refine the approach iteratively. We followed these steps:

\begin{itemize}
    \item Reviewing outputs manually for misclassifications.
    \item Identifying patterns in errors and improving prompt wording.
    \item Adjusting examples to clarify distinctions between categories.
    \item Running multiple evaluations to check consistency.
\end{itemize}

\subsection{Evaluation Criteria}
The outputs from OpenAI and OpenDeep are compared based on accuracy, consistency, and interpretability. Key evaluation metrics include:
\begin{itemize}
    \item Agreement between models on sentence categorization.
    \item Consistency in identifying entities A and B.
    \item Human validation of categorization correctness.
\end{itemize}

\section{Results and Discussion}

\subsection{Entity Extraction From Prompt Output}
When processing text from GPT-4o and DeepSeek R1, we found many issues with extra formatting and unnecessary text. These problems made cleaning and aligning the data challenging.

One issue in GPT-4o’s output was the frequent use of symbols like ***, ---, and **. These symbols served no useful purpose and appeared inconsistently. This made the outputs harder to parse. DeepSeek R1 often retained inline references and citations, including author names in parentheses. These additions were not always necessary for classification tasks.

Fluff text was another recurring issue. GPT-4o often added phrases such as:
"Here is the classification of sentences based on predefined categories."
These phrases were irrelevant and needed removal.

Formatting inconsistencies between the two models created further challenges. GPT-4o sometimes numbered its outputs, while DeepSeek R1 provided responses in a compact format. Both models presented key elements like categories, attributes (A and B), and sentences differently. This inconsistency made it difficult to directly compare their outputs.

To address these problems, we created a structured parsing framework. It uses regex-based rules to identify and extract sentences, categories, and attributes. The framework removes irrelevant text and normalizes formatting. Structured JSON files were prepared from the raw outputs of both models.

For alignment, fuzzy matching techniques were used. These techniques matched sentences between GPT-4o and DeepSeek R1, even when there were minor differences in phrasing. This approach improved the accuracy of mapping and ensured consistency in the processed data. The result was a clean and reliable dataset for analysis.

\subsection{Overall Categorization Coverage}

We analyzed the total number of sentences processed and the extent to which GPT-4o and DeepSeek R1 provided category assignments. Table~\ref{tab:overall_coverage} summarizes the categorization coverage across all analyzed sentences.

\begin{table}[h]
    \centering
    \begin{tabular}{l c c}
        \toprule
        & \textbf{ChatGPT} & \textbf{DeepSeek} \\
        \midrule
        Total Sentences & 1823 & 1823 \\
        Categorized & 1654 & 1738 \\
        N/A (No Category Assigned) & 169 & 85 \\
        \bottomrule
    \end{tabular}
    \caption{Overall categorization coverage for GPT-4o and DeepSeek R1.}
    \label{tab:overall_coverage}
\end{table}

Both models assigned relationship categories to most sentences. However, GPT-4o failed to categorize 169 sentences, while DeepSeek R1 failed in 85 cases. The difference suggests variation in handling ambiguous or complex sentence structures. DeepSeek R1 provided more complete coverage, categorizing 84 more sentences than GPT-4o. This indicates that DeepSeek R1 attempts to assign relationships in cases where GPT-4o remains uncertain. Manual investigation indicates DeepSeek R1 is more capable of assigning categories to sentences containing mathematical symbols.

\subsection{Data Processing Issues in Entity Agreement Analysis}

We analyzed how well GPT-4o and DeepSeek R1 agreed in three aspects: relationship category assignment, Entity A extraction, and Entity B extraction. The results show large inconsistencies between these two models.

The agreement between the two models in the task of category classification is 44.71\%. This suggests that GPT-4o and DeepSeek R1 interpret sentence structures and semantic relationships differently. The disagreement likely results from how each model applies category definitions. Some categories may overlap, leading to inconsistent choices.

Entity extraction shows a different pattern. The models agree on Entity A in 37.36\% of cases. This suggests that both models can identify the main subject of a sentence with moderate consistency. However, agreement on Entity B is lower at 22.44\%. The lower accuracy in Entity B extraction suggests more ambiguity. The models may struggle with implicit entities, multi-clause sentences, or domain-specific terms.

The large differences in classification and entity extraction show fundamental variations in how these models process scientific text. A deeper analysis is required to find the specific cases where misalignment occurs most often. This can help improve prompt design and make entity extraction more reliable.

\subsection{Category Agreement Between GPT-4o and DeepSeek R1}

\begin{figure}[h]
    \centering
    \includegraphics[width=0.95\linewidth]{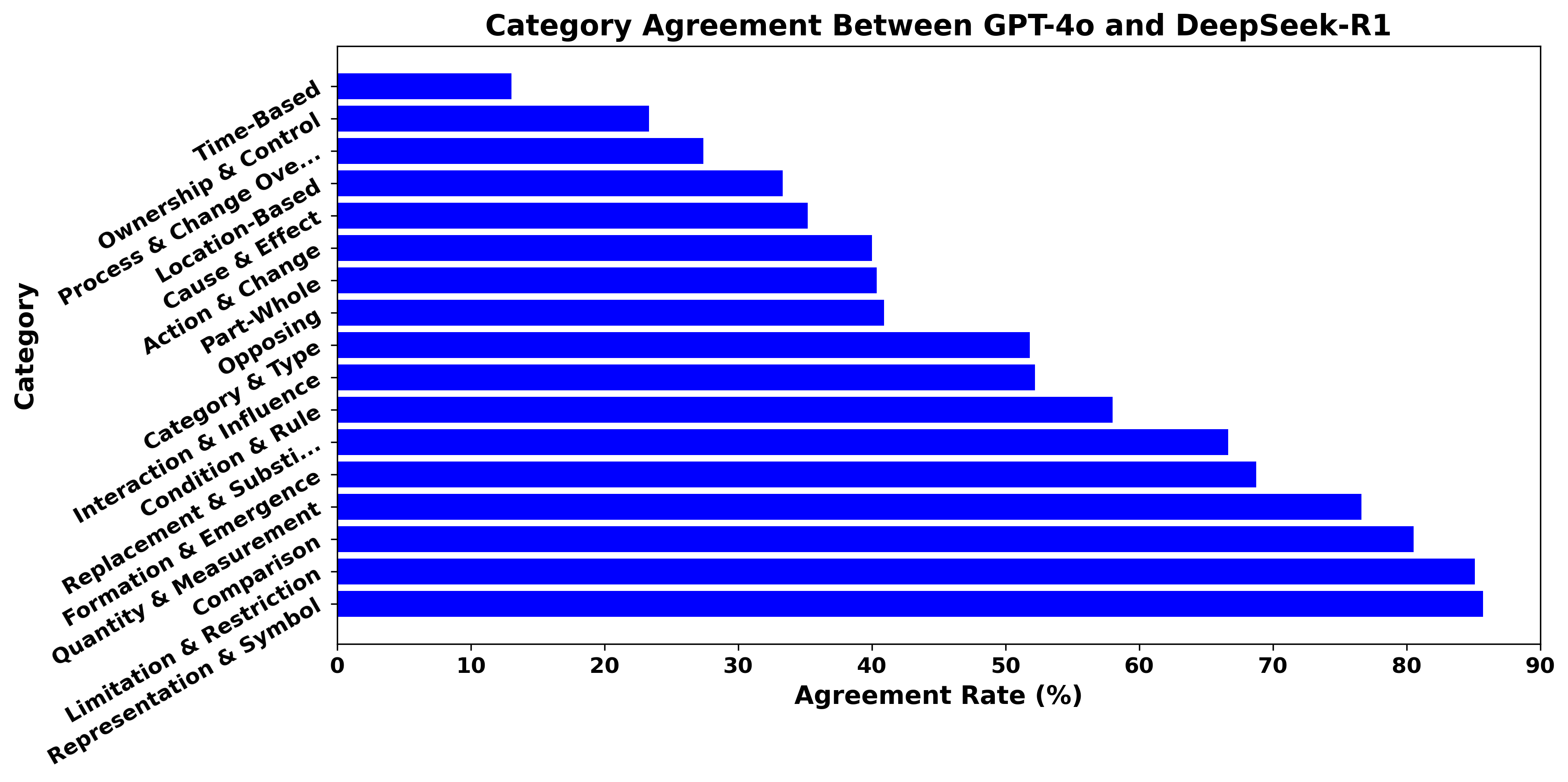}
    \caption{Agreement rates for different relationship categories assigned by GPT-4o and DeepSeek R1.}
    \label{fig:category_agreement}
\end{figure}

To assess how consistently GPT-4o and DeepSeek R1 classify scientific sentences, we analyzed their agreement on assigned relationship categories. The agreement rate for each category is shown in Fig.~\ref{fig:category_agreement}. Only the categories with nonzero agreement rates are shown.

GPT-4o and DeepSeek R1 show an overall category agreement of 44.71\%, with significant variation across categories. The highest agreement occurs in \textit{Representation \& Symbol Relationship (85.71\%)} and \textit{Limitation \& Restriction Relationship (85.11\%)}, indicating strong consistency in these categories. In contrast, \textit{Cause \& Effect Relationship (35.20\%)}, \textit{Ownership \& Control Relationship (23.33\%)}, and \textit{Time-Based Relationship (13.04\%)} exhibit low agreement, suggesting different heuristics in classification.

Misalignment is common in cases involving causal and temporal relationships. GPT-4o frequently assigns \textit{Interaction \& Influence Relationship}, while DeepSeek R1 tends to classify similar cases as \textit{Formation \& Emergence Relationship}. Additionally, several categories, including \textit{Function \& Purpose Relationship} and \textit{Mathematical Relationship}, show no agreement, meaning at least one model rarely uses them. The results suggest that GPT-4o and DeepSeek R1 apply different strategies when categorizing scientific text, leading to inconsistencies in classification.

\begin{figure}[H]
    \centering
    \includegraphics[width=0.85\textwidth]{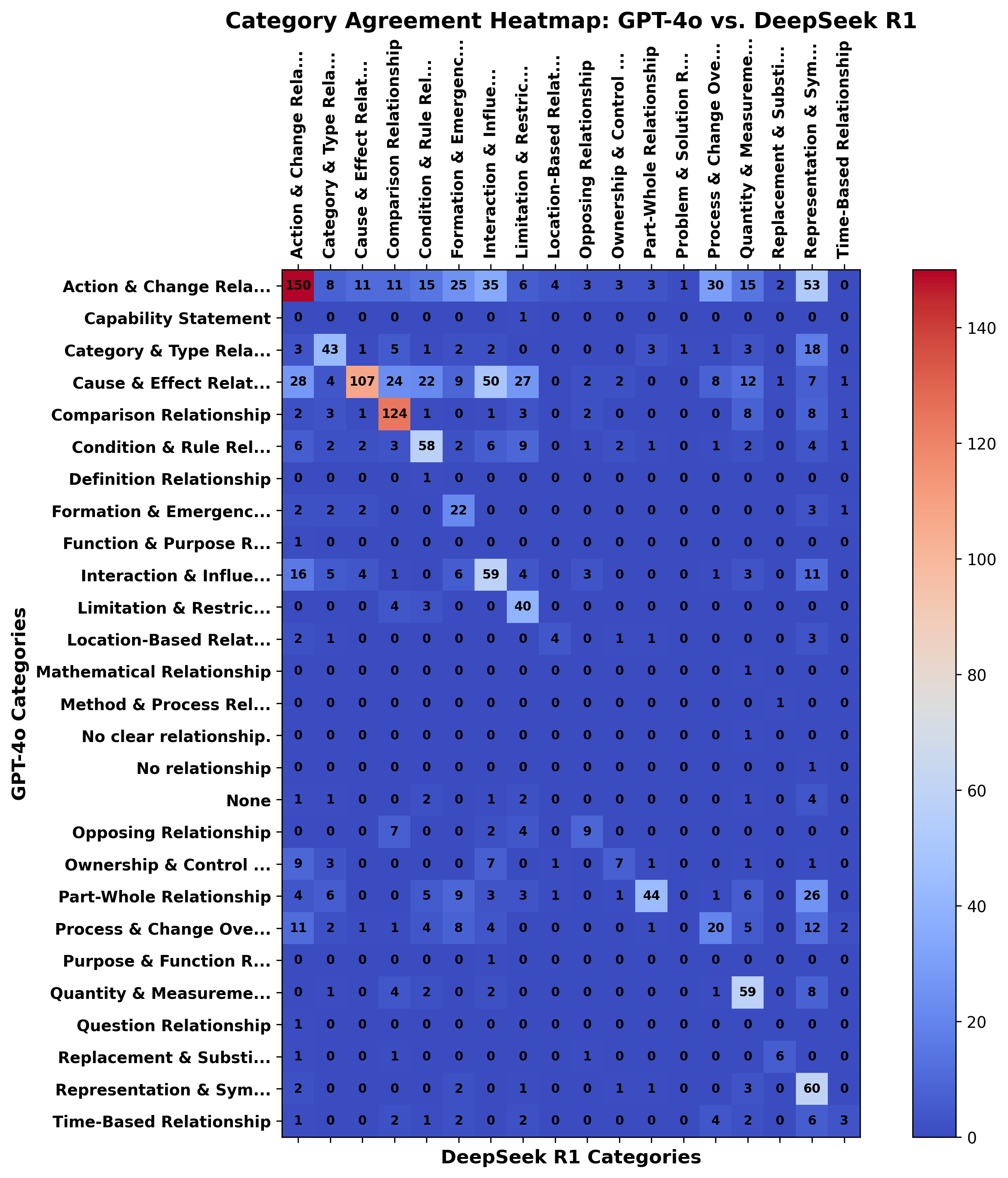}
    \caption{Pairwise agreement heatmap between GPT-4o and DeepSeek R1 category assignments. The heatmap shows how frequently both models assigned the same category to the same sentence.}
    \label{fig:category_agreement_heatmap}
\end{figure}

To show detailed pairwise agreement counts between GPT-4o and DeepSeek R1, we show a heatmap in Figure \ref{fig:category_agreement_heatmap}. The heatmap provides a structured view of how frequently both models assigned the same category to the same sentence. It shows areas of high agreement, as well as categories where misclassification is more common, offering deeper insights into discrepancies between the two models. In some cases, the model response processing code fails to capture the correct category due to inconsistencies in the model outputs. In such cases, we assign $None$. GPT-4o tends to classify causality broadly under \emph{Cause \& Effect}. DeepSeek R1, however, sometimes labels the same cases as \emph{Interaction \& Influence}, possibly treating causal relationships as mutual interactions.

\subsection{Entity Agreement Across Relationship Categories}

\begin{figure}[h]
   \centering
   \includegraphics[width=0.95\linewidth]{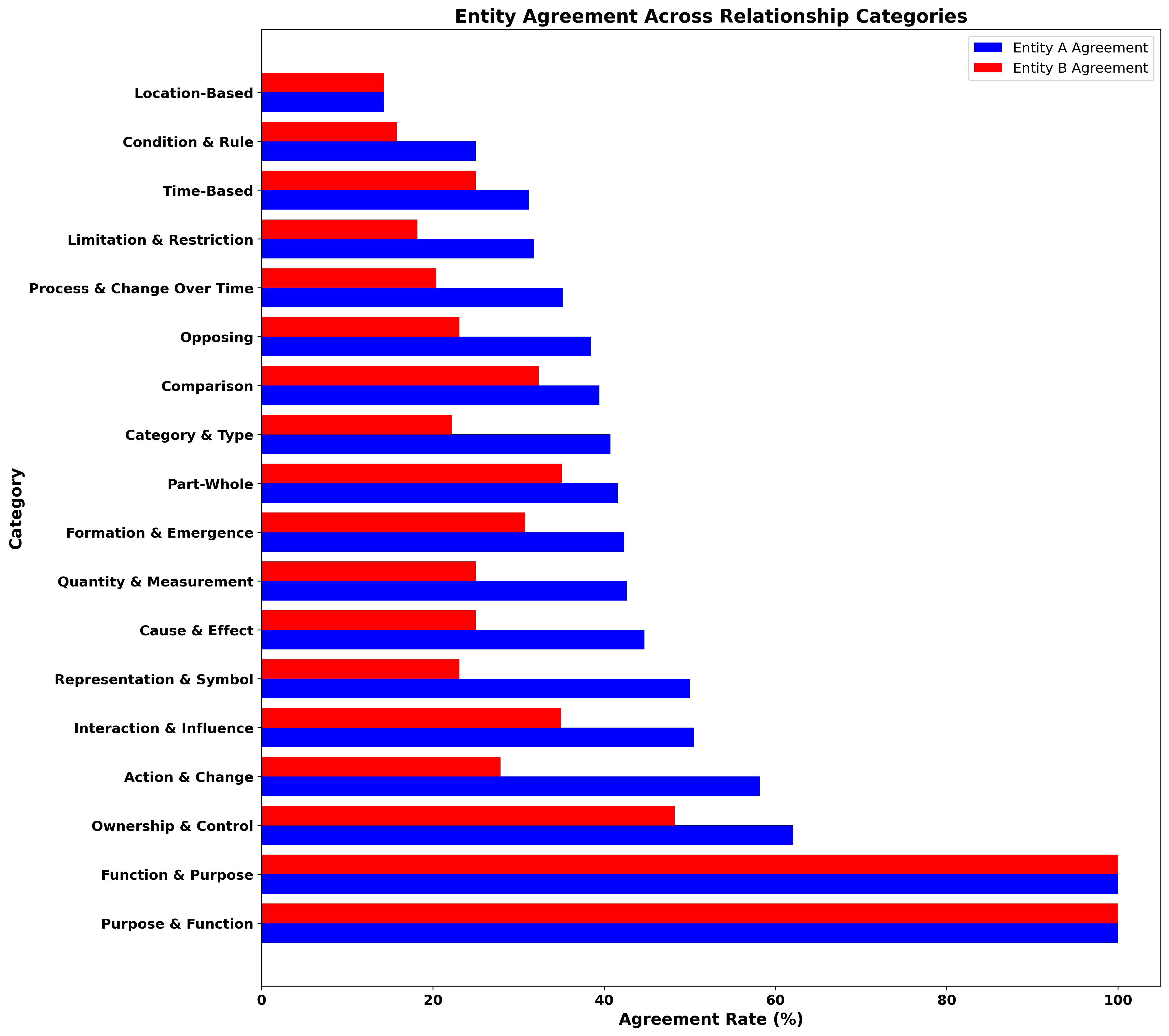}
   \caption{Entity agreement rates for different relationship categories between GPT-4o and DeepSeek R1.}
   \label{fig:entity_agreement}
\end{figure}

As shown in Figure~\ref{fig:entity_agreement}, agreement rates for Entity A are generally higher than those for Entity B. Only categories with nonzero agreement rates are plotted. Some categories, such as \textit{Function \& Purpose} and \textit{Purpose \& Function}, exhibit complete agreement for both entities, while others, including \textit{Formation \& Emergence} (46.88\% for Entity A, 31.25\% for Entity B) and \textit{Action \& Change} (42.13\% for Entity A, 19.47\% for Entity B), show mixed consistency. 

Entity B agreement remains lower across most categories, with several having near-zero agreement. The overall agreement rates across categories are 37.83\% for Entity A and 19.08\% for Entity B. This discrepancy suggests that the models interpret contextual dependencies differently, affecting their consistency in relational extraction. 

\subsection{Misclassification Patterns and Common Discrepancies}

GPT-4o and DeepSeek R1 often assign different categories to the same sentence. Table~\ref{tab:misclassification} shows an example.

\begin{table}[h]
    \centering
    \small
    \begin{tabular}{p{6cm} p{3cm} p{3cm}}
        \toprule
        \textbf{Sentence} & \textbf{GPT-4o} & \textbf{DeepSeek R1} \\
        \midrule
        We used data gathered during and after COVID-19 & Cause \& Effect & Interaction \& Influence \\
        \bottomrule
    \end{tabular}
    \caption{Example of a misclassified sentences.}
    \label{tab:misclassification}
\end{table}

Some patterns emerge on manual inspection of the dataset. \textit{Cause \& Effect} is often confused with \textit{Interaction \& Influence}. \textit{Action \& Change} sometimes overlaps with \textit{Condition \& Rule} or \textit{Formation \& Emergence}. Measurement-related sentences show mismatches between \textit{Action \& Change} and \textit{Quantity \& Measurement}. In some cases, GPT-4o assigns "N/A" where DeepSeek R1 detects a relationship. These inconsistencies indicate that DeepSeek R1 considers structural relationships which GPT-4o often does not.

\subsection{Overall Comparative Performance and Observations}
Our manual inspection reveals that \textit{ChatGPT} lacks consistency. It frequently misclassifies \textit{Action \& Change} instead of \textit{Interaction \& Influence}. Some labels miss key details, like performance limitations in cost-related discussions. The same relationship is sometimes assigned differently in similar contexts. \textit{DeepSeek R1} produces structured output but has unclear sentence boundaries. It often applies broad categories, such as \textit{Category \& Type} instead of \textit{Part-Whole}. Physical properties like bandwidth are sometimes misclassified under \textit{Quantity \& Measurement}.

\section{Conclusion and Future Work}

This study provides an early comparison of GPT-4o and DeepSeek R1 in categorizing sentences from scientific text. Given that DeepSeek R1 was released only a few weeks ago, its performance in specialized tasks remains largely unexplored. Our analysis is based on a limited dataset, and while it provides initial insights, a broader study with an expanded dataset is necessary to draw more definitive conclusions. 

Human evaluation was conducted in a limited manner, primarily to verify major discrepancies. A more comprehensive evaluation, involving multiple experts assigning scores to model responses, would provide a deeper understanding of strengths and weaknesses. Additionally, the predefined relationship categories used for classification may benefit from further refinement. Some categories showed frequent misclassification, suggesting that adjustments or hierarchical structuring could improve consistency. 

Future work will focus on refining the evaluation methodology. For instance, scaling the study to include a wider range of scientific texts, and improving the categorization framework to enhance model agreement can be explored as future direction.

%\section{Acknowledgment}
%This research was supported NOBODY - WE SPEND OUR OWN MONEY TO BUY APP :)
\bibliographystyle{IEEEtran}
\bibliography{references}

\end{document}